\documentclass[letterpaper, 10 pt, conference]{ieeeconf}
\IEEEoverridecommandlockouts
\usepackage[colorlinks]{hyperref}
\hypersetup{
    colorlinks,
    linkcolor=black,
    citecolor=black,
    filecolor=magenta,
    urlcolor=cyan,
}

\usepackage{resizegather}
\usepackage{todonotes}
\usepackage{cite}
\usepackage{amsmath,amssymb,amsfonts,mathrsfs}
\usepackage{graphicx}
\usepackage{textcomp}
\usepackage{xcolor}
\usepackage{tcolorbox}

\usepackage{tablefootnote}
\usepackage{threeparttable}
\usepackage{caption, subcaption}
\usepackage{bm}
\usepackage{tikz}
\usepackage{graphicx}
\usepackage{tkz-euclide}

\usepackage{algorithm}
\usepackage[noend]{algpseudocode}
\usepackage{algorithmicx}

\usetikzlibrary{positioning}
\usetikzlibrary{shapes}
\usetikzlibrary{shapes.misc}
\usetikzlibrary{shapes.geometric}
\usetikzlibrary{plotmarks}
\usetikzlibrary{intersections}
\usetikzlibrary{calc}
\usetikzlibrary{fit}
\usetikzlibrary{patterns,tikzmark}
\usetikzlibrary{matrix,decorations.pathreplacing,calc}

\tikzset{cross/.style={cross out, draw, 
         minimum size=2*(#1-\pgflinewidth), 
         inner sep=0pt, outer sep=0pt}}
\newcommand{\vertiii}[1]{{\left\vert\kern-0.25ex\left\vert\kern-0.25ex\left\vert #1 
    \right\vert\kern-0.25ex\right\vert\kern-0.25ex\right\vert}}

\makeatletter
\renewcommand{\fps@figure}{htp}
\renewcommand{\fps@table}{htp}
\makeatother

\def\BibTeX{{\rm B\kern-.05em{\sc i\kern-.025em b}\kern-.08em
    T\kern-.1667em\lower.7ex\hbox{E}\kern-.125emX}}
    
\begin{document}

\title{BiRoDiff: Diffusion policies for bipedal robot locomotion on unseen terrains}

\author{GVS Mothish, Manan Tayal, Shishir Kolathaya
\thanks{
This work is supported by Stochastic Robotics Lab at IISc,
}
\thanks{Cyber-Physical System, Indian Institute of Science (IISc), Bengaluru.
{\tt\scriptsize \{mothishg, manantayal, shishirk\}@iisc.ac.in}}}%

\maketitle
\begin{abstract}
% Control of legged robots in unknown terrains is a difficult task, especially when the robot is not exposed to these types during training. 
 Locomotion on unknown terrains is essential for bipedal robots to handle novel real-world challenges, thus expanding their utility in disaster response and exploration. In this work, we introduce a lightweight framework that learns a single walking controller that yields locomotion on multiple terrains. We have designed a real-time robot controller based on diffusion models, which not only captures multiple behaviours with different velocities in a single policy but also generalizes well for unseen terrains. Our controller learns with offline data, which is better than online learning in aspects like scalability, simplicity in training scheme etc. We have designed and implemented a diffusion model-based policy controller in simulation on our custom-made Bipedal Robot model named Stoch BiRo. We have demonstrated its generalization capability and high frequency control step generation relative to typical generative models, which require huge onboarding compute.
\end{abstract}

% \begin{IEEEkeywords}
%     Keywords : Bipedal Robots, Diffusion Models, Latent Space, Embeddings, Learning based Controller, Neural Networks.
% \end{IEEEkeywords}

\section{Introduction}
\label{section: Introduction}
\par Legged locomotion is a highly effective way of navigating many types of environments that are designed for humans. Navigating complex and varied terrains is a fundamental challenge in robotics, particularly for legged systems. Bipedal locomotion, which must adapt to both seen and unseen environments, demands a high degree of agility and adaptability.  Learning-based methods for robot control, particularly those utilizing Deep Reinforcement Learning (DRL), as explored in studies like \cite{heess2017emergence,rudin2022learning}, have demonstrated significant progress in mapping observations or robot states to actions through policy learning. Some approaches, such as those in \cite{li2021reinforcement,kumar2022adapting}, parameterize the policy and value functions, which quantify the state and state-action values, respectively, and learn these parameters. Works like \cite{paigwar2021robust,castillo2022reinforcement, tayalrealising,9682564} have successfully implemented learned policies to achieve quadrupedal and bipedal locomotion. However, these methods often require a lot of samples to train good policies, moreover, they train different policies for different terrains.

 % \todo{NEED TO Transition better here. You ended the previous paragraph saying single policy. But next paragraph is about BC?}
In order to improve the sample efficiency of training, there is a need to learn the walking behaviours from demonstrations. Behaviour cloning \cite{torabi2018behavioral} learns the behaviour in a supervised fashion from the offline demonstrations, but it causes problems like covariance shift. To tackle this problem, methods like DAgger\cite{daggerross2011reduction}, apprenticeship learning \cite{abbeel2004apprenticeship}, offline RL \cite{levine2020offline,xu2022policy, kumar2020conservative} were introduced. But in practice, challenges such as the existence of multimodal distributions, sequential correlation, and the requirement of high precision make this task distinct and challenging compared to other supervised learning problems. Previous works such as \cite{mandlekar2021matters} used mixture of Gaussians to solve these kind of problems. Recent advances in Generative models, which are very effective in capturing multimodal distributions and learning distributions, have also inspired many to apply them to robotics problems. 

Following the huge success of VAEs, GANs and other generative models, Diffusion models (inspired by non-equilibrium thermodynamics) have shown promising applications in planning and control problems. Planning with Diffusion models is first proposed in \cite{janner2022planning}. They have modelled trajectories as state-action pairs of multiple time steps, and DDPM \cite{ho2020denoising} is used to model the distribution. Following this,  \cite{ajay2022conditional} modelled the problem into two parts; in the first part, the diffusion model learns the state-only trajectory and in the second, it learns inverse dynamics to obtain actions. Many of the other works built upon this were SafeDiffuser \cite{xiao2023safediffuser}, which aims to safe generation in data by applying contraints. Another work \cite{liang2023adaptdiffuser} formulated a rule-based method for improving trajectories. Policy learning in robotics is another approach in which diffusion models are showing prominent results. SfBC \cite{chen2022offline} imitates behaviour policy in an Inverse Reinforcement Learning setting. Additionally, IDQL \cite{huang2024diffuseloco} designed the sampling based on Implicit Q-Learning Framework, Diffusion-Ql encourages to minimise the Q-function difference. 
 
 While Decision Diffuser \cite{ajay2022conditional} and  DiffuseLoco \cite{huang2024diffuseloco} are the works that applied diffusion models to legged robots. Decision Diffuser demonstrated the advantages of modelling policies as conditional diffusion models for gait generation in quadrupedal robots, which is a two-stage process of learning model using diffusion models and an Inverse dynamics model for obtaining actions. DiffuseLoco used Transformers with DDPMs to diffusion models to learn directly from offline multimodal datasets with a diverse set of locomotion skills. However, these methods are computationally intensive and not suitable for real-time implementation on legged robots with limited computational resources. We have developed a real-time diffusion model based controller for bipedal walking in unseen terrains called BiRoDiffuser with a learnable latent space.

The main contributions of our work are as follows: 

\begin{enumerate}
    \item We have developed a novel lightweight, multi-terrain controller that can learn agile bipedal walking in a single policy.
    \item Our framework, which utilizes a diffusion model, learns to generate walking behaviours on the terrains, that the robot is not trained on.
    \item We validate the efficacy of the framework on the Isaac Gym environment of our custom bipedal robot Stoch-BiRo.
    
\end{enumerate}

\begin{figure}[h]
\centering
\begin{subfigure}[b]{0.45\textwidth}
  \centering
  \includegraphics[width=\textwidth]{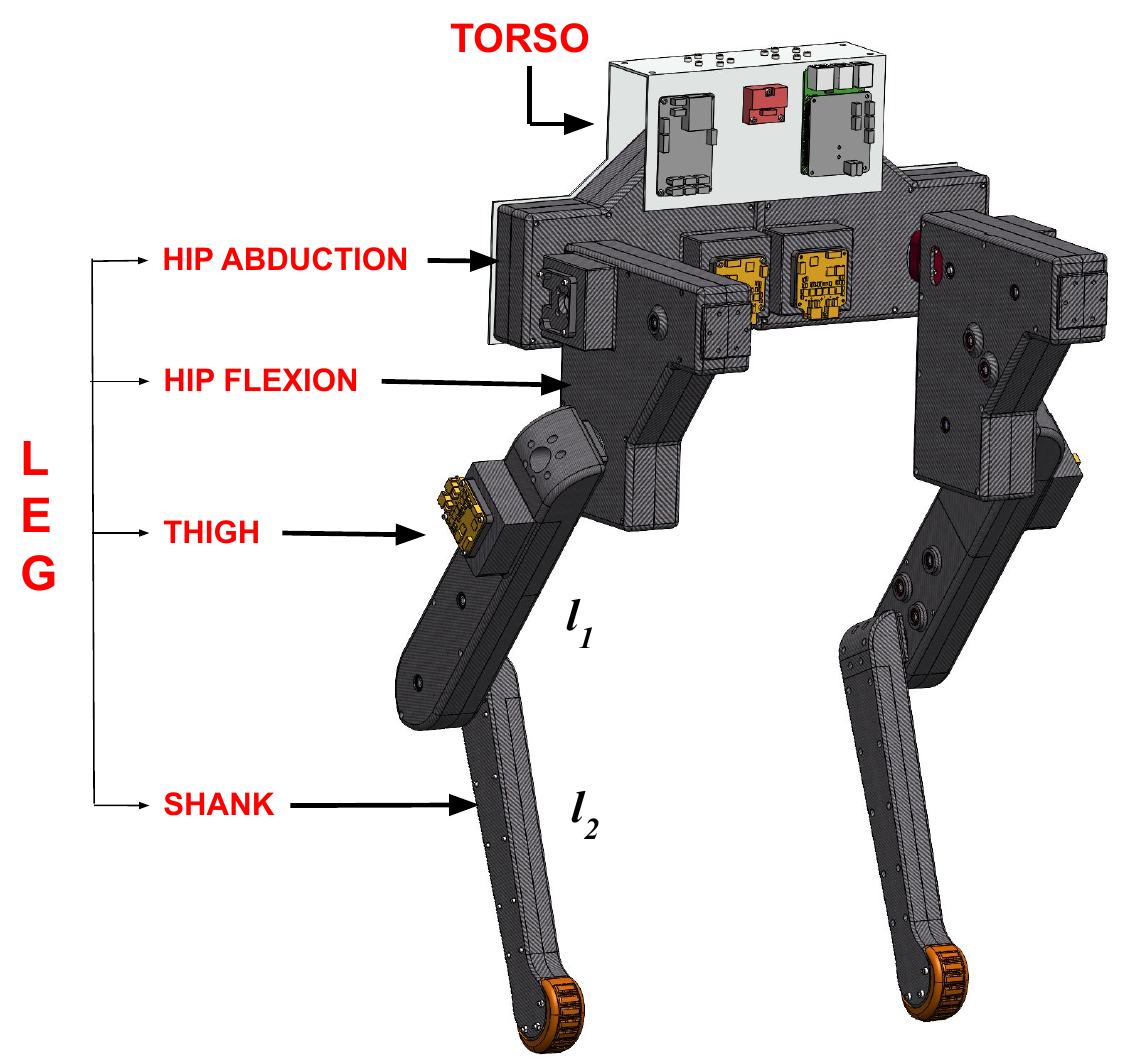}
\end{subfigure}
% \begin{subfigure}[b]{0.23\textwidth}
% \centering
%     \includegraphics[width=\textwidth]{images/Hardware/Master_Assembly_Exploded.pdf}
%     \caption{Exploded View}
%     \label{exploded}
% \end{subfigure}
\caption{Stoch BiRo}
\label{fig: stoch-biro}
\end{figure}
\subsection{Organisation}

First, a detailed background of required concepts used in this work is briefly explained in Section \ref{section: Background}. The data collection procedure for offline learning is described in Section \ref{section: Data}. Our controller architecture and diffusion network details are explained in Section \ref{section: Diff controller}. The simulation results are provided in Section \ref{section: Results}. At last, we present our conclusion in Section \ref{section: Conclusions}.

\section{Background}
\label{section: Background}
This section will give some background on the description of our custom bipedal robot Stoch-BiRo, followed by Diffusion Probabilistic Models and finally, diffusion models in robotics or Diffusers.

\subsection{Description of Stoch-BiRo}

Our custom-made bipedal robot, Stoch BiRo \cite{mothish2024stoch} is a low-cost and modular robot with a point foot. It has six controllable degrees of freedom (3 DOF per leg) and is a simple robot designed to navigate unstructured and unseen terrains. The torso is the base link in which both the legs are connected through revolute joints. Each leg has three links, Hip, Thigh, and Shank, connected through two joints. This robot (Fig. \ref{fig: stoch-biro}) is both position-controlled and torque-controlled. We are using position control, which is preferred for our controller as learning joint angles in action space is easier than learning joint torques because of smoothness in sequence of actions.

\subsection{Diffusion Probabilistic Models}
Diffusion Probabilistic Models slowly destroy the data distribution by adding noise iteratively and learn to denoise the data from noise.
Diffusion Probabilistic Models pose the data-generating process as an iterative denoising procedure $ p_\theta(\mathbf{x}_{t-1} \vert \mathbf{x}_t) $. This denoising is the reverse of a forward diffusion process $q(x_{t}|x_{t-1})$ that slowly corrupts the structure in data by adding noise. The data distribution induced by the model is given by:
\begin{equation}
    p_\theta(\mathbf{x}_{0:T}) = p(\mathbf{x}_T) \prod^T_{t=1} p_\theta(\mathbf{x}_{t-1} \vert \mathbf{x}_t) \\p_\theta(\mathbf{x}_{t-1} \vert \mathbf{x}_t) = \mathcal{N}(\mathbf{x}_{t-1}; \boldsymbol{\mu}_\theta(\mathbf{x}_t, t), \boldsymbol{\Sigma}_\theta(\mathbf{x}_t, t))
\end{equation}
The reverse process is often parameterized as Gaussian with fixed timestep-dependent covariances.
where $p(\mathbf{x}_T)$ is a standard Gaussian prior and $\mathbf{x}_{0}$ denotes (noiseless) data. Parameters $\theta$ are optimized by minimizing a variational bound on the negative log-likelihood of the reverse process.

\subsection{Diffusers}
The work \cite{janner2022planning} is one of the earliest works that used diffusion models in robotics. They have proposed a method called Diffuser, which is based on Model-based Reinforcement Learning, as it learns to de-noise a 2-dimensional array (Trajectory) of both state and actions 
\begin{equation}
    \mathcal{T} = 
    \begin{bmatrix}
s_{0} & s_{1} & . & . & . &  s_{T} \\ 
a_{0} & a_{1} & . & . & . &  a_{T}
\end{bmatrix}
\end{equation}
simultaneously using the diffusion model, this helps to establish a stronger relation between modeling and planning. To predict a single time step, they used a concept of receptive field, which constrains the model and enforces local consistency. In contrast to typical trajectory planners, their idea is to simultaneously predict all the time steps. The core idea behind combining multiple steps is that local consistency can ensure global coherence in a generated plan.
They have used a guide function $\mathcal{J}$, which helps in optimizing the test-time objective or satisfying a set of constraints

\vspace{10pt}

\textbf{Notations :} In the rest of the paper, we use subscripts to denote information on control time steps and superscripts to denote information on diffusion time steps.

\section{Data Collection}
\label{section: Data}
We have implemented a deep reinforcement learning policy using an established work by \cite{rudin2022learning} in legged locomotion. The training configuration is designed to generate policies for real-world robotic tasks, and learned using Proximal Policy Optimization (PPO) \cite{schulman2017proximal} through extensive parallel processing on a single workstation GPU. Following the training and deployment of the policy on our robot, we collected the data for the training a generalised policy using the framework described in the section \ref{section: Diff controller}. The data $\mathcal{D}$ comprises of observation-action pairs $\mathcal{D} = \{(\mathbf{O}_i, \mathbf{a}_i)\}_{i=1}^N$, where N represents the number of data points, which is 500500 in this instance. This includes 250250 pairs for each velocities 0.3 m/s and 1 m/s, encompassing walking on flat ground as well as slopes. We collected equal amounts of walking data from slopes with inclinations of 5.7 and 10.2 degrees. Each observation has a size of 150, representing the states of the past five control time steps to predict the next step. each state (which has a size of 30) consists of base linear velocities, base angular velocities, projected gravity vector, commands, the difference in actuator position from the default position, actuator velocity, and previous actions selected by the policy. Actions consists of the joint angles of 6 actuators. 

% There is a PD controller that tracks the commanded joint angles.

\begin{table}[h]
        \centering
        \caption{Data of behaviours collected for training}
        \begin{tabular}{c c c }
        \hline
            Velocity  & Flat Ground & Slopes\\
            \hline
             1 m/s & $\checkmark$ & 5.7$^{\circ}$ and 10.2$^{\circ}$\\
             0.3 m/s & $\checkmark$ & 5.7$^{\circ}$ and 10.2$^{\circ}$\\
        \hline
        \end{tabular}
        \label{table: diff_data}
    \end{table}

\begin{figure}[h]
    \centering
    \begin{subfigure}[b]{0.48\linewidth}
        \centering
        \includegraphics[width=\linewidth]{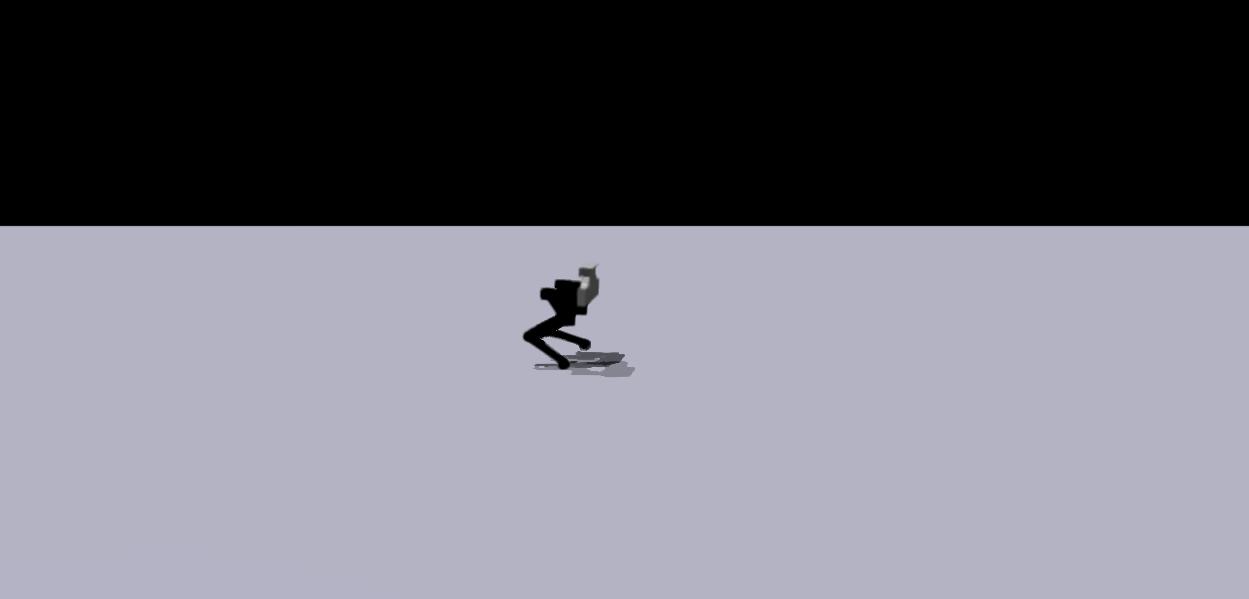}
        \caption{Flat Ground}
        \label{fig: flat_data}
    \end{subfigure}
    \begin{subfigure}[b]{0.48\linewidth}
        \centering
        \includegraphics[width=\linewidth]{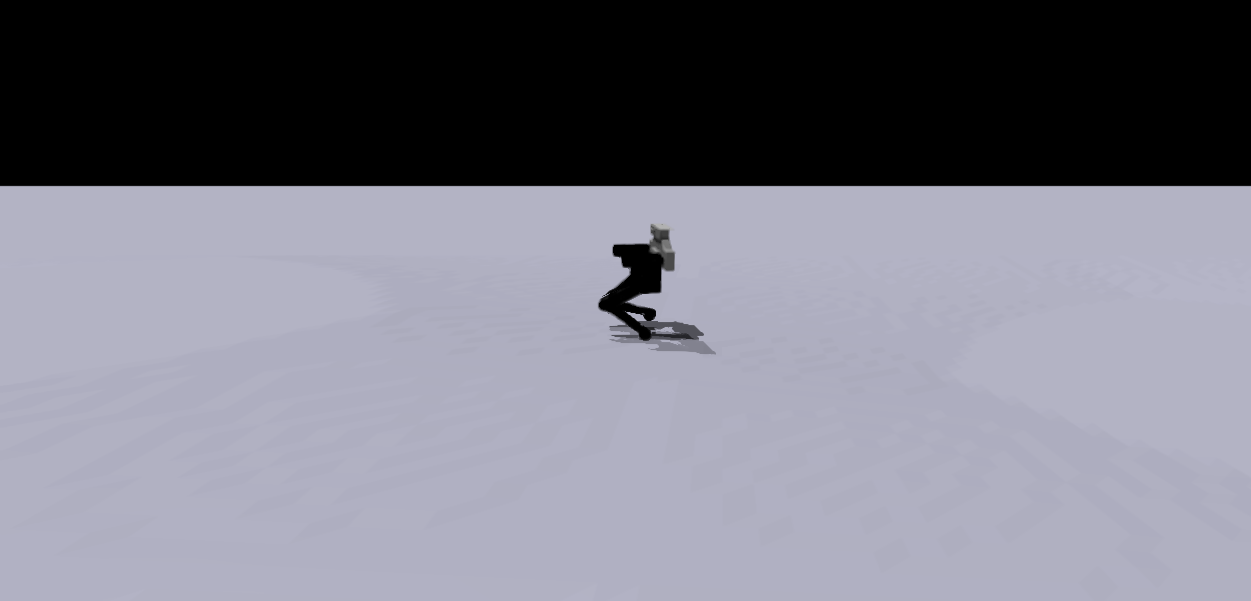}
        \caption{Slopes}
        \label{fig: slopes_data}
    \end{subfigure}
    \caption{Biped walking data collected on (a) flat ground and (b) slopes to be used for training the diffusion policy}
    \label{fig: terrains_data}
\end{figure}

A point to be noted is that although the source policies used to collect the dataset in this work are RL-based policies, our framework is general and can include the data generated from model-based optimal controllers (such as from \cite{tayal2023safe}). The requirement is to align the state and action spaces among different source policies, and the frequency of the policy should be kept the same.

\section{Diffusion Policy}
\label{section: Diff controller}
We have developed a model in which actions are denoised from noise using a diffusion model conditioned on latent observations. We have used the DDPM (Denoising Diffusion Probabilistic Models)\cite{ho2020denoising} model for the diffusion process.

 \begin{figure*}[h]
    \centering
    \includegraphics[width=1.0\linewidth, trim={0 120 0 150}, clip]{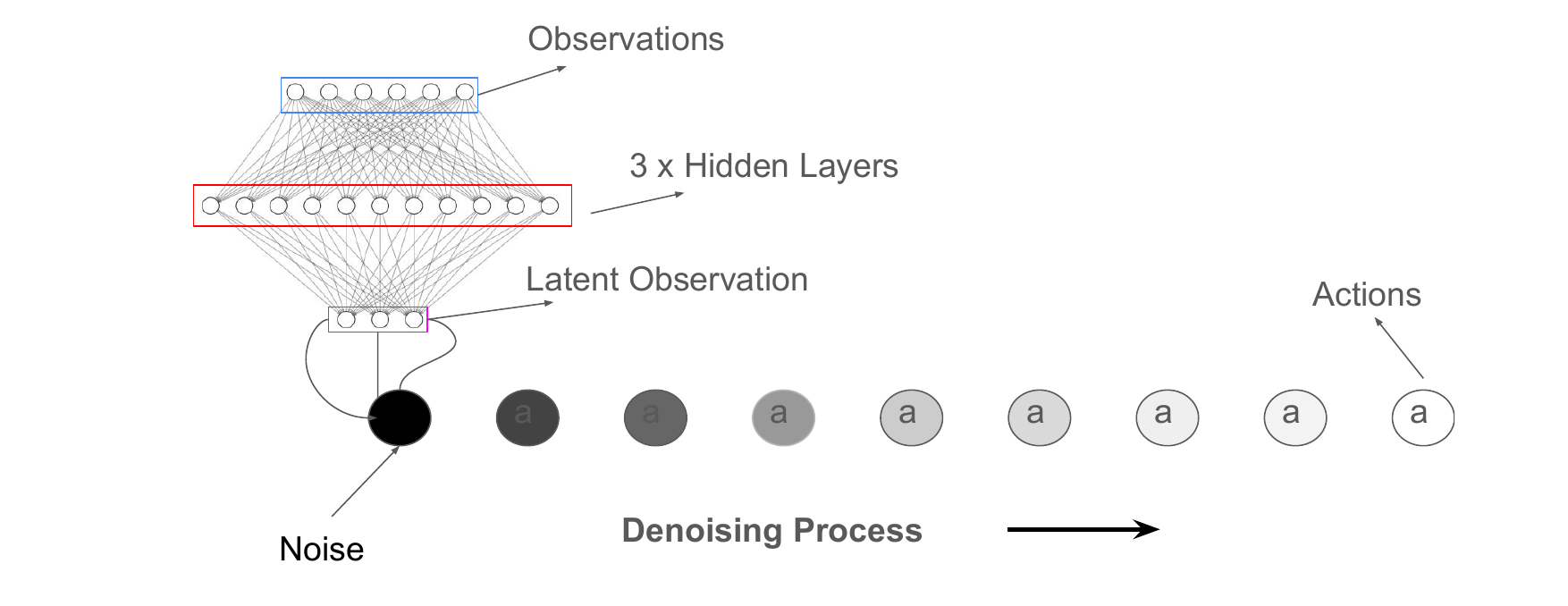}
\caption{Architecture of Diffusion Policy : (a)The Neural Network $\mathcal{M}$ which takes input as observations, having three hidden layers outputs the latent observations. (b) The diffusion process is represented as actions being denoised in multiple steps. A network $\bm{\epsilon_{\theta}}$ outputs the noise, which has to be separated in each step.}
\label{diffusion_archi}
\end{figure*}

\subsection{Diffusion Policy Controller Architecture}
 At first, observations are passed through an MLP, which we refer to as $\mathcal{M}$ and the output of the network can be seen as latent observations $\mathbf{O}^{l}$ since this network has learnable parameters. These observations, concatenated with noise and time step embeddings $\bm{t_{emb}}$, which use sinusoidal embeddings of size 128, are input to the diffusion network. The diffusion network $\bm{\epsilon_{\theta}}$ denoises the actions from noise, conditioned on embeddings, which contain time embeddings and latent observations.
 \begin{equation}
     \bm{t_{emb}} = Sinusoidal Embedding (t)
 \end{equation}
\begin{equation}
    \textbf{emb} = \{\bm{t_{emb}},\mathbf{O}_{t_{ctrl}}^{\mathbf{l}}\}
\end{equation}
\begin{equation}
 \mathbf{a}_{t_{ctrl}}^{t-1} \leftarrow \mathbf{a}_{t_{ctrl}}^{t} - \bm{\epsilon_{\theta}}(\mathbf{a}_{t_{ctrl}}^{t}, \textbf{emb}))
\end{equation}
 
 as shown in Fig. \ref{diffusion_archi}. We have used linear noise scheduling for betas, 60 denoising time steps.

 Our latent observation network $\mathcal{M}$ is a 3-layer MLP with GELU activation function \cite{gelu}. The denoising network is also an MLP with 7 layers and a GELU activation function. We prefer using an MLP because of its lightweight and faster sampling rate in the reverse process, which results in real-time control frequency for action generation from the policy. The GELU activation function gives the effect of both RELU and Dropout, which is generally used as an activation function in generative models. Refer to table \ref{diff_params} for more details.

\begin{table}[h]
    \centering
    \caption{Diffusion Policy Architecture Parameters}
    \begin{tabular}{c c c}
    \hline
        Parameter & Value\\
        
        \hline
           &  \\
         Size of latent Space Observation & 48\\
         No of observations & 30*5(4 steps history) = 150\\
         No of actions & 6 angles (6 DOF)\\
         Learning rate & 1e-4\\
         Hidden dimensions of latent Network & 48\\
         Hidden dimensions of Denoising Network & 256\\
         No of hidden layers in the Latent network  & 3\\
         No of hidden layers in the Denoising network & 7\\
         No of epochs & 10000\\
         Batch Size & 4000\\
         Time embedding size & 128\\
         Time embedding type & Sinusoidal embedding\\
         Beta Schedule & linear\\
         No of denoising Steps & 60\\
         No of obs-action pairs for training & 500500\\
    \hline
    \end{tabular}
    \label{diff_params}
\end{table}

\subsection{Training Procedure}
We trained the diffusion policy using the data $\mathcal{D}$ collected as explained in Section \ref{section: Data}. Observations are input to the network which outputs the denoised actions and MSE loss (Mean square error) 
\begin{equation*}
    \mathcal{L} = mse loss(\bm{\epsilon},\bm{\epsilon_{\theta}}(\mathbf{a}^{t}, \textbf{emb}))
\end{equation*}
is calculated between denoised actions and actions from the data. ADAM \cite{kingma2014adam} optimizer is used to optimize the loss function and learn the parameters of both networks. Since both the networks are connected, backpropagation is performed to learn the parameters of both networks simultaneously, as shown in Algorithm \ref{algorithm_training}.
\begin{algorithm}[H]
    \caption{Diffusion Policy Training}
    \label{algorithm_training}
    \begin{algorithmic}[1]
        \Require Offline Dataset $\mathcal{D} = \{(\mathbf{O}_i, \mathbf{a}_i)\}_{i=1}^N$

        \For {epoch in num epochs}
            \For{no of training Steps}
                \State Sample batch $\mathcal{B} = \{\mathbf{O},\mathbf{a}\}$ from $\mathcal{D}$            \State observations $\mathbf{O}$, actions $\mathbf{a}^0$ $\leftarrow$ $\mathcal{B}$
                \State Sample Noise $\bm{\epsilon} \sim \bm{\mathcal{N}}(0, 1)$
                \State t $\sim$ Uniform({1,...,T})
                \State noisy actions $\mathbf{a}^{t} =  \sqrt{\bm{\bar{\alpha_t}}}\mathbf{a}^{0} + \sqrt{1 - \bm{\bar{\alpha_t}}}\bm{\epsilon}$
                \State Predict Latent Observations $\mathbf{O}^{\mathbf{l}} = \mathcal{M} \left (  \mathbf{O} \right )$
                \State $\bm{t_{emb}}$ = Sinusoidal Embedding (t)
                \State Embeddings \textbf{emb} = \{$\bm{t_{emb}},\mathbf{O}^{\mathbf{l}}$\}
                % \State noise predicted  = $\epsilon_{\theta}(\mathbf{a}^{t}, emb)$
                \State Loss $\mathcal{L} = mse loss(\bm{\epsilon}$,$\bm{\epsilon_{\theta}}(\mathbf{a}^{t}, \textbf{emb})$)
                \State optimize $\theta$ and parameters of $\mathcal{M}$ using $\mathcal{L}$
            \EndFor
            \State Save the model in every 1000 epochs
        \EndFor
    \end{algorithmic}
\end{algorithm}

During sampling, both the models are exported to the simulation environment and for every control time step, action is sampled from the policy conditioned on observations. This action from the policy is applied to the environment, and the next observation is used to sample the next action as shown in Algorithm \ref{algorithm_sampling}.
\begin{algorithm}[H]
    \caption{Diffusion Policy Sampling}
    \label{algorithm_sampling}
    \begin{algorithmic}[1]
        \Require $\mathcal{M}$, $\bm{\epsilon}_{\theta}$
        \State $\mathbf{O}_{0} \leftarrow $ (Initial Observation ($t_{ctrl} = 0$))
        \State Sample $\mathbf{a}_{t_{ctrl}}^{T} \sim \mathcal{N}(0, \bm{I})$
        
        % \State $\mathbf{O}_{0}^{\mathbf{l}} \leftarrow \mathcal{M}\left ( \mathbf{O}_{0} \right )$
        \While{alive}
            \State $\mathbf{O}_{t_{ctrl}}^{\mathbf{l}} \leftarrow \mathcal{M}\left ( \mathbf{O}_{t_{ctrl}} \right )$
            \For{$t = T$ to $1$}
                \State $\bm{t_{emb}}$ = Sinusoidal Embedding (t)
                \State \textbf{emb} = \{$\bm{t_{emb}},\mathbf{O}_{t_{ctrl}}^{\mathbf{l}}$\}
                \State $\mathbf{a}_{t_{ctrl}}^{t-1} \leftarrow \mathbf{a}_{t_{ctrl}}^{t} - \bm{\epsilon_{\theta}}(\mathbf{a}_{t_{ctrl}}^{t}, \textbf{emb})$)
            \EndFor
            \State $\mathbf{O}_{t{ctrl+1}} \leftarrow$ Apply action $\mathbf{a}_{t_ctrl}^{0}$ on Robot
            \State $t_{ctrl} \leftarrow t_{ctrl} + 1$
        \EndWhile
    \end{algorithmic}
\end{algorithm}

\section{Results}
\label{section: Results}
We present our simulation results using NVIDIA’s Isaac Gym simulation environment \cite{makoviychuk2021isaac}. The training was conducted on an AMD Ryzen Threadripper PRO 5975WX CPU paired with an Nvidia GeForce RTX 4090 GPU, resulting in a total training time of 2 hours for the diffusion policy. We used an 80-20 split for training and validation sets, achieving a mean squared error (MSE) of 0.00874 for the training data and 0.0101 for the validation data. The loss exhibited an exponential decrease during the initial epochs, followed by a gradual decrement.

Initially, we focused on learning known behaviors such as walking on flat grounds and slopes using our diffusion policy. Following successful evaluation on these terrains, we extended the evaluation to unknown surfaces that were not included in the training data, such as rough terrain, rough slopes, steps, and discrete terrain. The behavior for each terrain is detailed in the following subsections.

\subsection{Flat Ground}
We trained the model with flat ground walking data for 50 seconds at both velocities of 1 m/s and 0.3 m/s. The diffusion policy successfully learned to walk on flat ground for more than 60 seconds at both velocities as shown in Fig. \ref{fig: resuts_behaviour}.

\subsection{Slopes}
The training data included discrete slopes of 5.7 and 10.2 degrees at both velocities of 0.3 m/s and 1 m/s. During sampling from the diffusion policy, the robot managed to walk on slopes up to 18.4 degrees at 1 m/s and up to 16.7 degrees at 0.3 m/s for more than 60 seconds. This indicates that the diffusion policy can interpolate walking behaviors for slopes not included in the training data, as well as handle slightly steeper slopes. This can be seen in Fig. \ref{fig: resuts_behaviour}.

\subsection{Rough Terrain}
No rough terrain walking data was used during training. Nevertheless, the policy generated actions that enabled walking on rough terrain for 20 seconds at 1 m/s and more than 60 seconds at 0.3 m/s as shown in Fig. \ref{fig: resuts_behaviour}.

\subsection{Rough Slopes and Steps}
The diffusion policy was not specifically trained on rough slopes or steps. Despite this, the policy successfully generated walking behaviors on these terrains at a velocity of 1 m/s as shown in Fig. \ref{fig: resuts_behaviour}.

\subsection{Discrete Terrain}
\begin{figure}
        \centering
        \begin{subfigure}{\linewidth}
        \includegraphics[width=1.0\linewidth]{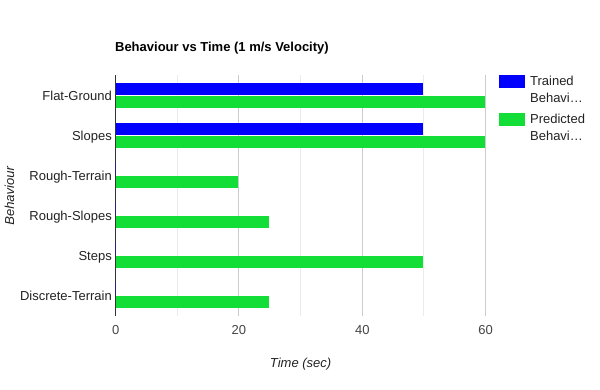}
            % \caption{Velocity : 1 m/s}
            \label{fig: 1ms}
            \caption{}
        \end{subfigure}
        \begin{subfigure}{\linewidth}
        \includegraphics[width=1.0\linewidth]{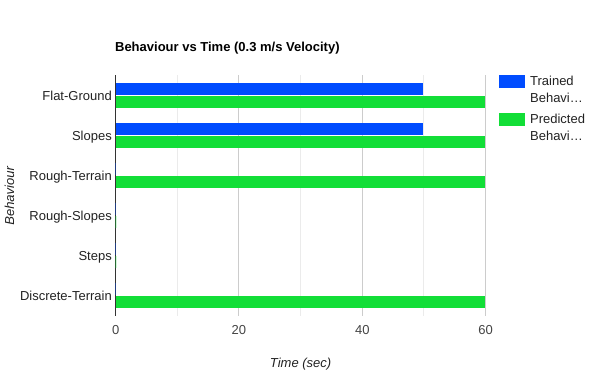}
            % \caption{Velocity : 0.3 m/s}
            \label{fig: 0_3ms}
            \caption{}
        \end{subfigure}
        \hfill
    \caption{Walking behaviour on different terrains}
    \label{fig: resuts_behaviour}
\end{figure}

The diffusion policy generated walking trajectories for 30 seconds on discrete terrain at 1 m/s and more than 60 seconds at 0.3 m/s as shown in Fig. \ref{fig: resuts_behaviour}.

\begin{figure}
    \centering
    \begin{subfigure}[b]{0.8\linewidth}
        \centering
        \includegraphics[width=\linewidth]{images/new_image_sim/flat.png}
        \caption{Flat Ground}
        \label{flat}
    \end{subfigure}
    \begin{subfigure}[b]{0.8\linewidth}
        \centering
        \includegraphics[width=\linewidth]{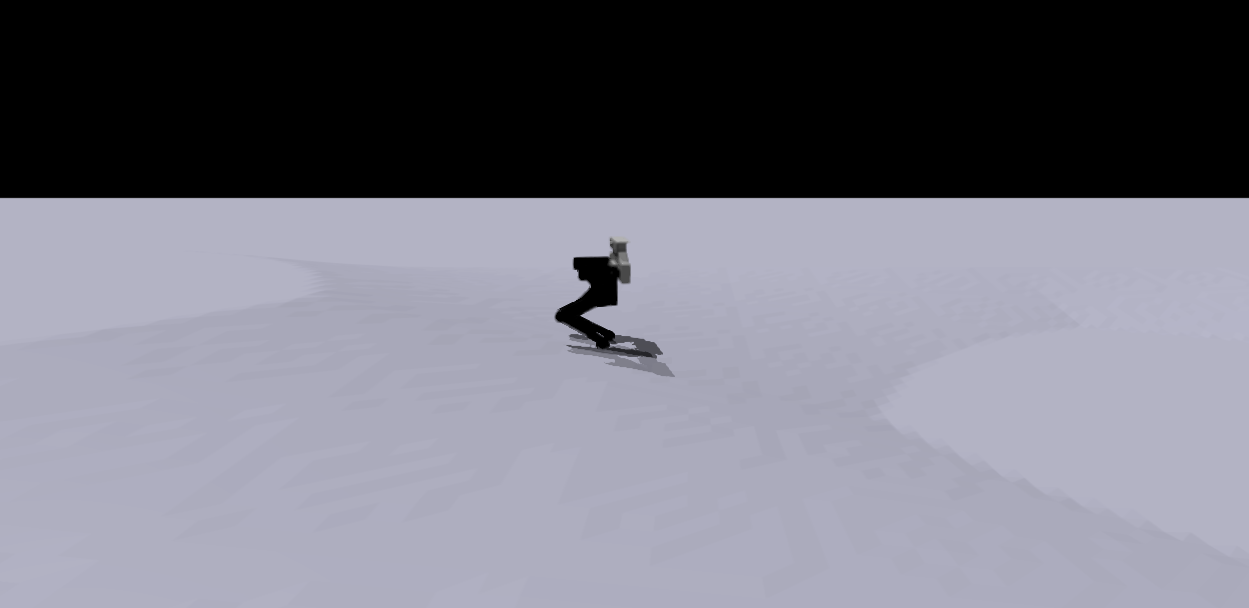}
        \caption{Slopes}
        \label{slopes}
    \end{subfigure}
    \begin{subfigure}[b]{0.8\linewidth}
        \centering
        \includegraphics[width=\linewidth]{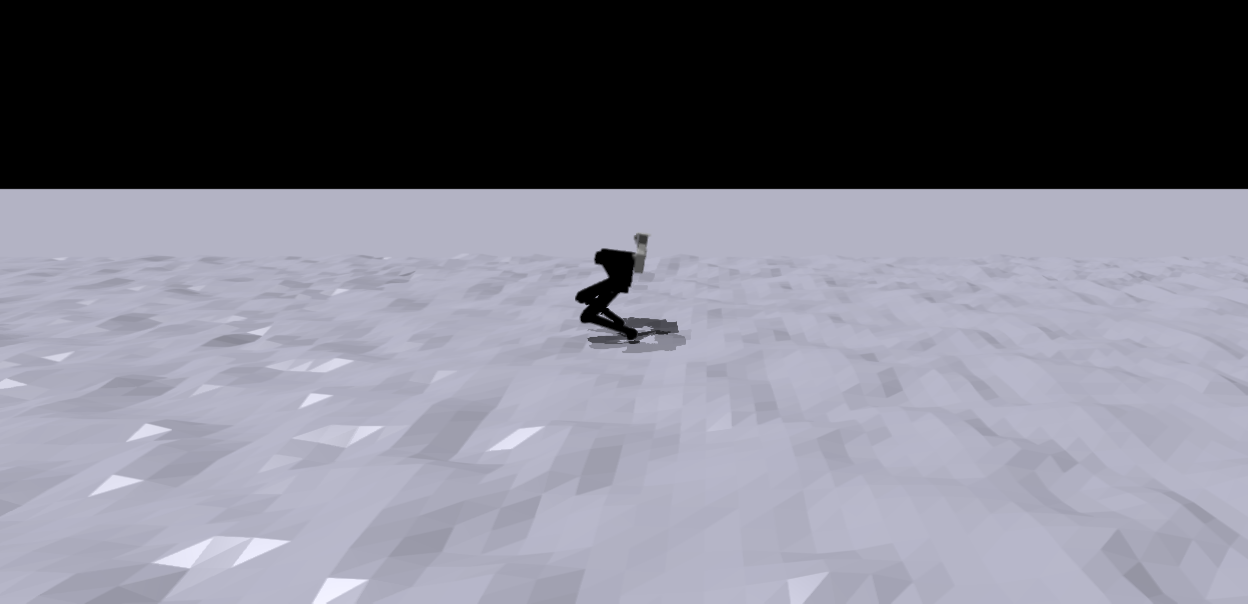}
        \caption{Rough Terrain}
        \label{rough}
    \end{subfigure}
    \begin{subfigure}[b]{0.80\linewidth}
        \centering
        \includegraphics[width=\linewidth]{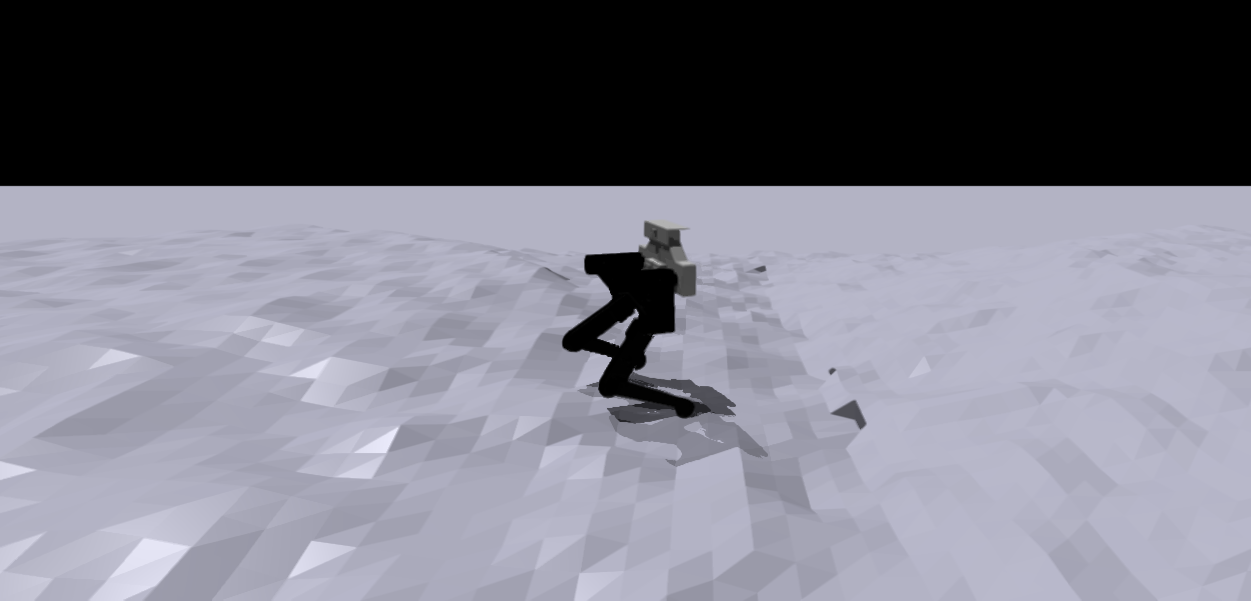}
        \caption{Rough Slopes}
        \label{roughslopes}
    \end{subfigure}
    \begin{subfigure}[b]{0.8\linewidth}
        \centering
        \includegraphics[width=\linewidth]{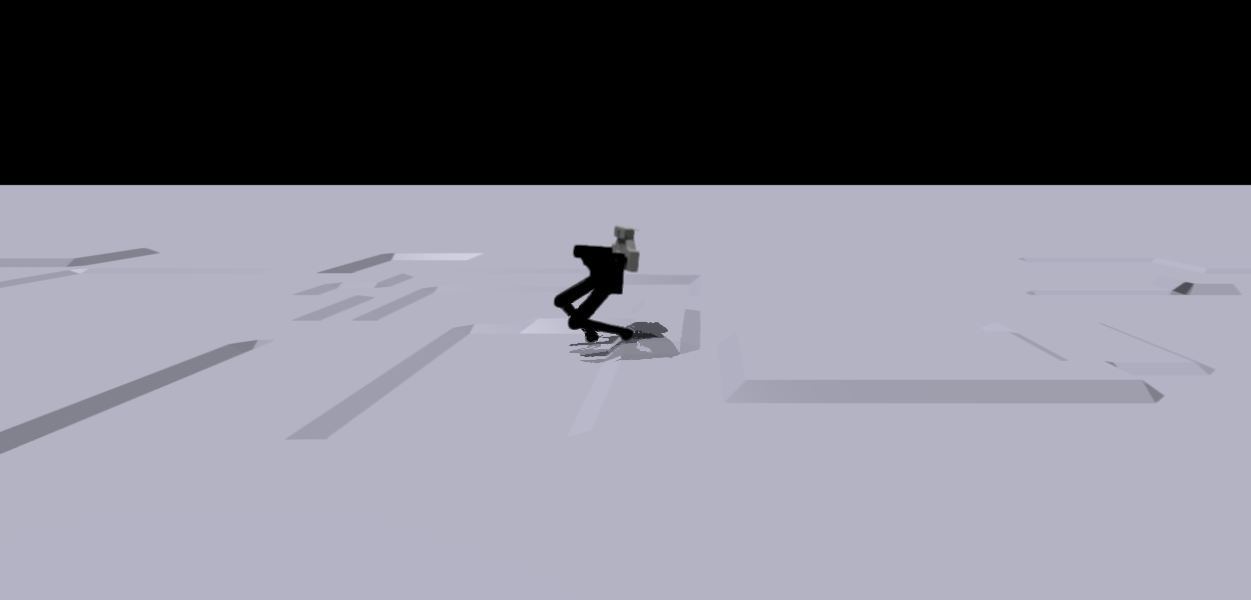}
        \caption{Discrete Terrain}
        \label{discrete}
    \end{subfigure}
    \begin{subfigure}[b]{0.8\linewidth}
        \centering
        \includegraphics[width=\linewidth]{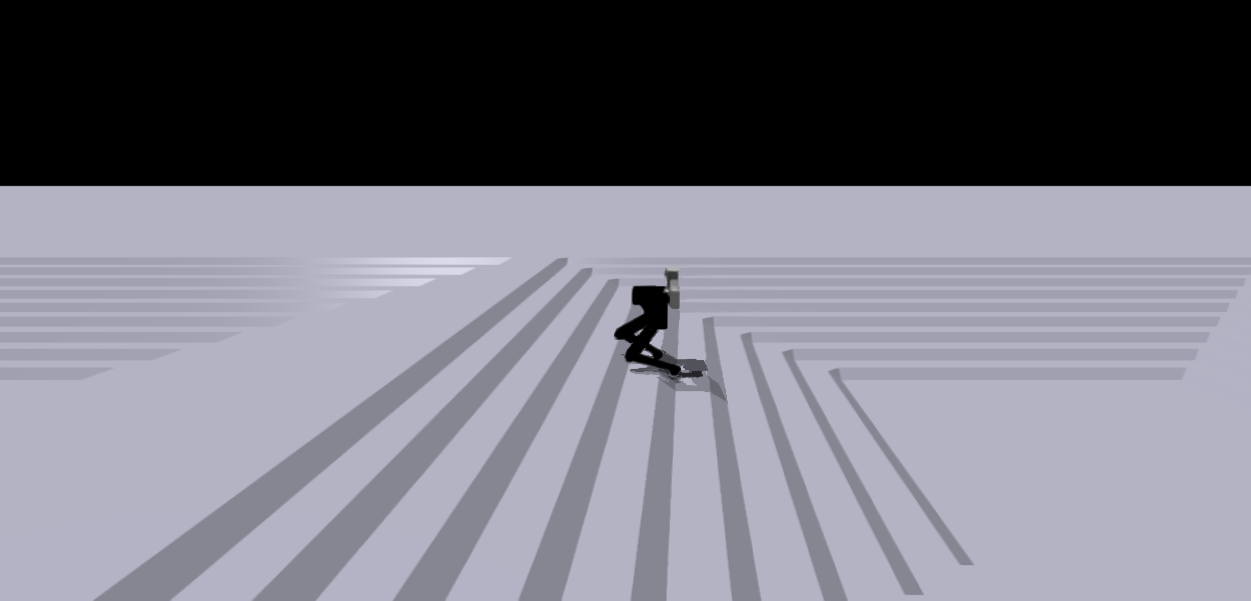}
        \caption{Steps}
        \label{steps}
    \end{subfigure}
    \caption{Stoch-BiRo navigating across different unseen terrains using the BiRoDiff policy}
    \label{resuts_terrains}
\end{figure}

Fig \ref{resuts_terrains} shows the walking behaviours of the bipedal robot in different unseen terrains. The supplementary video showing these walking behaviours is available on our website\footnote{\label{icc_webpage}\url{https://gvsmothish.github.io/Stoch-Biro-diffusion-policy/}}.

We used the joint angles of the robot as action space for our controller, the other approach is to use joint torques. During our experiments, we realized that diffusion in joint angle space is easier than torque space since angle trajectories are more smoother than torques, where sudden changes are evident in sequential torque values.

\begin{table*}[h]
        \centering
        \caption{Behavioural Results}
        \begin{tabular}{c c c c c c c}
        \hline
            Velocity  & Flat Ground & Slopes & Rough Terrain & Rough Slopes & Steps & Discrete Terrain\\
            \hline
            % outlook.office365.com/mail/\vspace{0.3cm}
             1 m/s & $\checkmark$ & upto 18.4$^{\circ}$ & $\checkmark$ (20sec) & upto 13.0$^{\circ}$ (25sec)& $\checkmark$ & $\checkmark$ (25sec)\\
             0.3 m/s & $\checkmark$ & upto 16.7$^{\circ}$ & $\checkmark$ & x & x & $\checkmark$\\
        \hline
        \end{tabular}
        \label{diff_results}
    \end{table*}

\section{Conclusions \& Discussion}
\label{section: Conclusions}
A Diffusion based policy controller is designed and deployed in the Simulation, which is very effective and powerful in learning Multiple behaviours and Interpolating Behaviours. As we have demonstrated in Section \ref{diff_results}, we have achieved walking on known terrains and on unknown terrains. It is evident that the robot is falling after some time in uneven terrains, but this can be solved by training with more velocities because the results signify that velocity plays a major role in balancing unseen terrains. Velocity tracking of the controller is not as expected and can be improved by training the network for more time steps and with more data. This controller is limited by Short-sightedness as it only generates a single future time step from the past time steps.

\subsection{Discussions and Future Work}
(1) The diffusion policy can be improved by predicting more future time steps instead of a single step and, in parallel, achieving sampling time which is necessary for good control of the robot. (2) Deploying these diffusion policies in hardware and making these types of policies more realistic. (3) Integrating Image/video data with the architecture in latent space will uncover the greater potential of this kind of generative policies.

% \section{Acknowledgement}
% \input{acknowledgement}

\label{section: References}
\bibliographystyle{IEEEtran}
\bibliography{references.bib}

\end{document}